\documentclass[10pt,twocolumn,letterpaper]{article}
\pdfoutput=1
\usepackage{multirow} 
\usepackage{tcolorbox}
\usepackage{pifont}

\usepackage{colortbl}
\usepackage{marvosym}
\usepackage[pagenumbers]{iccv} % To force page numbers, e.g. for an arXiv version

\newcommand{\PromptSty}[1]{\textnormal{\color{blue!90!black}#1}\unskip}

\definecolor{iccvblue}{rgb}{0.21,0.49,0.74}
\usepackage[pagebackref,breaklinks,colorlinks]{hyperref}
\definecolor{lowred}{RGB}{238,18,137}

\definecolor{lowerred}{RGB}{255,110,180}

\newcommand{\dplus}[1]{\fontsize{6pt}{0.1em}\selectfont (\textbf{\textcolor{lowred}{#1}})}

\makeatletter
\def\blfootnote{\xdef\@thefnmark{}\@footnotetext}
\makeatother
\def\paperID{xxx} % *** Enter the Paper ID here
\def\confName{ICCV}
\def\confYear{2025}

\title{ORION: A Holistic End-to-End Autonomous Driving Framework by Vision-Language Instructed Action Generation}

\author{Haoyu Fu$^{1*}$, Diankun Zhang$^{2*}$, Zongchuang Zhao$^{1*}$, Jianfeng Cui$^{2}$, Dingkang Liang$^{1\dag}$, \\ 
Chong Zhang$^{2}$, Dingyuan Zhang$^{1}$, Hongwei Xie$^{2\dag}$, Bing Wang$^{2}$, Xiang Bai$^{1}$\\
\\
$^{1}$ Huazhong University of Science and Technology, $^{2}$ Xiaomi EV\\
{\tt \{hyfu, zcuangzhao, dkliang\}@hust.edu.cn} \\
\url{https://xiaomi-mlab.github.io/Orion/}
}

\begin{document}
\maketitle
% \textsuperscript{\Letter}
\makeatletter\def\Hy@Warning#1{}\makeatother
\blfootnote{\noindent * Equal contribution. $\dag$ Project leader. Work done when Haoyu Fu and Zhongchuang Zhao were interns at Xiaomi EV.}

\begin{abstract}
End-to-end (E2E) autonomous driving methods still struggle to make correct decisions in interactive closed-loop evaluation due to limited causal reasoning capability. Current methods attempt to leverage the powerful understanding and reasoning abilities of Vision-Language Models (VLMs) to resolve this dilemma. However, the problem is still open that few VLMs for E2E methods perform well in the closed-loop evaluation due to the gap between the semantic reasoning space and the purely numerical trajectory output in the action space. To tackle this issue, we propose ORION, a h\textbf{O}listic E2E autonomous d\textbf{R}iving framework by v\textbf{I}sion-language instructed acti\textbf{ON} generation. ORION uniquely combines a QT-Former to aggregate long-term history context, a Large Language Model (LLM) for driving scenario reasoning, and a generative planner for precision trajectory prediction. ORION further aligns the reasoning space and the action space to implement a unified E2E optimization for both visual question-answering (VQA) and planning tasks. Our method achieves an impressive closed-loop performance of 77.74 Driving Score (DS) and 54.62\% Success Rate (SR) on the challenge Bench2Drive datasets, which outperforms state-of-the-art (SOTA) methods by a large margin of 14.28 DS and 19.61\% SR. 

\end{abstract}

%%%%%%%%%%%%%%%%%%%%%%%%%%%%%%%%%%%%%
%%%%%%%  Introduction
%%%%%%%%%%%%%%%%%%%%%%%%%%%%%%%%%%%%%%

\begin{figure}[tp!]
    \centering
    \includegraphics[width=0.47\textwidth]{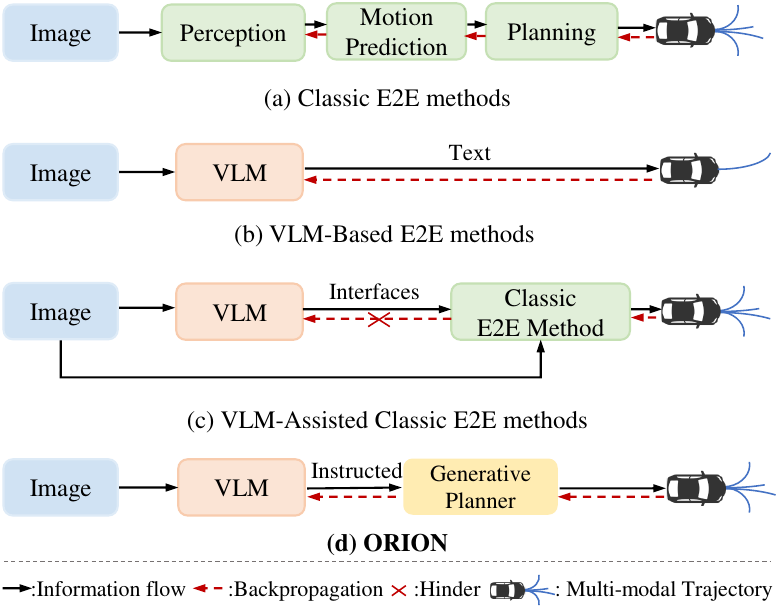}
    \caption{The comparison of different E2E paradigms. Our ORION framework establishes the differentiable connection between reasoning and action space via the generative planner.}
    \label{fig: introduction}
    % \vspace{-10pt}
\end{figure}

\section{Introduction}
End-to-end (E2E) autonomous driving has witnessed significant advancements in recent years. Classic E2E methods~\cite{hu2023planning,jiang2023vad, zhang2024sparsead, zheng2024genad, chen2024vadv2} integrate perception~\cite{philion2020lift,jiang2024far3d,zhang2023fully}, prediction~\cite{gu2023vip3d,shi2022motion,chai2019multipath}, and planning~\cite{hu2022st, prakash2021multi} modules through multi-task learning, as shown in Fig.~\ref{fig: introduction}(a). These methods optimize driving trajectories by imitating expert demonstrations, achieving promising performance in the open-loop evaluation~\cite{caesar2020nuscenes,sun2020scalability}. Nevertheless, these methods lack the common sense to complete complex causal reasoning. As a result, they struggle with comprehensive closed-loop benchmarks~\cite{jia2024bench2drive} that require autonomous decision-making and dynamic environmental interactions. Recently, Vision-Language Models (VLMs)~\cite{lu2024deepseek, openai2023gpt, chen2024internvl,wang2024qwen2vl} have accumulated rich world knowledge and aligned vision-language space through large-scale data training, providing new insight for achieving E2E autonomous driving. 

Despite these advances, leveraging VLMs for E2E autonomous driving is not trivial, as the capabilities of VLMs focus on the semantic reasoning space, while E2E methods only need the numerical planning results in the action space. Some methods~\cite{wang2024omnidrive, hwang2024emma,wang2023drivemlm, xu2024drivegpt4,shao2024lmdrive} attempt to directly output text-based planning results using VLM, as shown in Fig.~\ref{fig: introduction}(b). Although this paradigm is convenient, VLM is not well-suited for handling mathematical calculations or numerical reasoning~\cite{frieder2024mathematical, peng2021mathbert}. Besides, limited by the intrinsic autoregressive mechanism of VLM, this framework only infers single results, which is inconsistent with the natural uncertainty of human planning~\cite{chen2024vadv2}. Therefore, directly using VLM for E2E autonomous driving may produce suboptimal solutions in complex scenes~\cite{xing2025openemma}. Other methods endeavor to bridge the gap via utilizing VLM output meta-action (\textit{e.g.}, turn left) to assist classic E2E methods~\cite{jiang2024senna, mei2024continuously}, as shown in Fig.~\ref{fig: introduction}(c). They adopt a carefully crafted interface to transmit the reasoning space information into the action space. However, this paradigm decouples these two spaces, hindering collaborative optimization between the trajectory optimization and the VLM reasoning process. Thus, the capabilities of VLM for E2E planning are not fully leveraged by the above framework.

To tackle this problem, we propose a h\textbf{O}listic E2E autonomous d\textbf{R}iving framework by v\textbf{I}sion-language instructed acti\textbf{ON} generation, termed ORION. Inspired by the field of conditional generation~\cite{kingma2013auto,liu2024sora, rombach2022high,ronneberger2015u}, where the semantic information controls the generation of detailed image features, we find that the generative model can construct a unified distribution of diverse types of data (\textit{e.g.}, image, text). Therefore, considering that the reasoning space of VLM and the action space of trajectory belong to different domains, we introduce a generative planner to establish a unified latent representation for aligning the two spaces. With the help of the introduced module, we take advantage of VLMs' reasoning information to construct trajectory, facilitating the model to capture the causal relationship between scene information and driving behavior.

Furthermore, it is well-known that long-term memory is necessary for E2E autonomous driving since historical information often influences trajectory planning within the current scene. Existing VLMs for E2E methods~\cite{hwang2024emma,xing2025openemma} typically concatenate multi-frame images for temporal modeling. They are constrained by the token length of VLM and incur significant computational overhead. Instead, motivated by OmniDrive~\cite{wang2024omnidrive}, which extracts features through Q-Former-styled architecture, we introduce QT-Former, a query-based temporal module. By leveraging a memory bank and a set of history queries, QT-Former effectively stores and extracts essential historical scene information to aggregate long-term visual context, further enhancing the temporal perception ability of reasoning and action space.

We evaluate the closed-loop driving ability of ORION on the Bench2Drive dataset, which builds interactive scenarios based on the CARLA~\cite{dosovitskiy2017carla} simulator. ORION achieves 77.74 Driving Score (DS) and 54.62\% Success Rate (SR), surpassing previous SOTA methods~\cite{jiadrivetransformer} with 14.28 driving scores and 19.61\% success rates, demonstrating the powerful superiority of ORION.

\textbf{The benefits of ORION are from three aspects:} 1) Thanks to the capability of the generative model to characterize the latent distribution of data, we bridge the gap between the reasoning space of VLM and the action space of trajectories through a generative planner, enabling the VLM to understand the scene and instruct trajectory generation.
2) The QT-former in ORION effectively captures long-term temporal dependencies, enabling the model to integrate temporal vision context into reasoning and action spaces. 3) Without bells and whistles, ORION achieves excellent performance in the Bench2Drive closed-loop benchmark. Experiments also show that ORION is compatible with diverse generative models, which further demonstrate the flexibility of our proposed framework.

%%%%%%%%%%%%%%%%%%%%%%%%%%%%%%%%%%%%%
%%%%%%%  Related Work
%%%%%%%%%%%%%%%%%%%%%%%%%%%%%%%%%%%%%%
\section{Related work}
\label{sec: Related work}

\subsection{End-to-End Autonomous Driving}
End-to-end autonomous driving~\cite{wu2022trajectoryguided, zhang2021roach} aims to directly process raw sensor data to predict motion trajectories or control signals, jointly optimizing the entire system to minimize error accumulation. Recent works like UniAD~\cite{hu2023planning} and VAD~\cite{jiang2023vad} integrate perception, prediction, and planning into a unified planning framework, making the framework ultimately planning-oriented. VADv2~\cite{chen2024vadv2} introduces probabilistic planning, outputting the probabilistic distribution of action and sampling one action to control the vehicle. GenAD~\cite{zheng2024genad} and DiffusionDrive~\cite{liao2024diffusiondrive} explore a new paradigm for end-to-end autonomous driving, employing the generative model to predict multi-modal trajectories. However, these methods mainly excel in open-loop evaluation, where the model could readily overfit to the ego status, as highlighted in Ego-MLP~\cite{zhai2023rethinking} and BEV-Planner~\cite{li2024ego}. Although some studies~\cite{chen2024vadv2, zheng2024genad, jia2023thinktwice, jia2023driveadapter} adopt closed-loop evaluation in CARLA~\cite{dosovitskiy2017carla} to assess robust driving ability, their performance remains suboptimal, revealing a notable gap between their open-loop and closed-loop results. Thus, we aim to construct an E2E autonomous driving system that demonstrates excellent performance in both open-loop and closed-loop evaluations.

\begin{figure*}[t]
    \centering
    \includegraphics[width=0.98\textwidth]{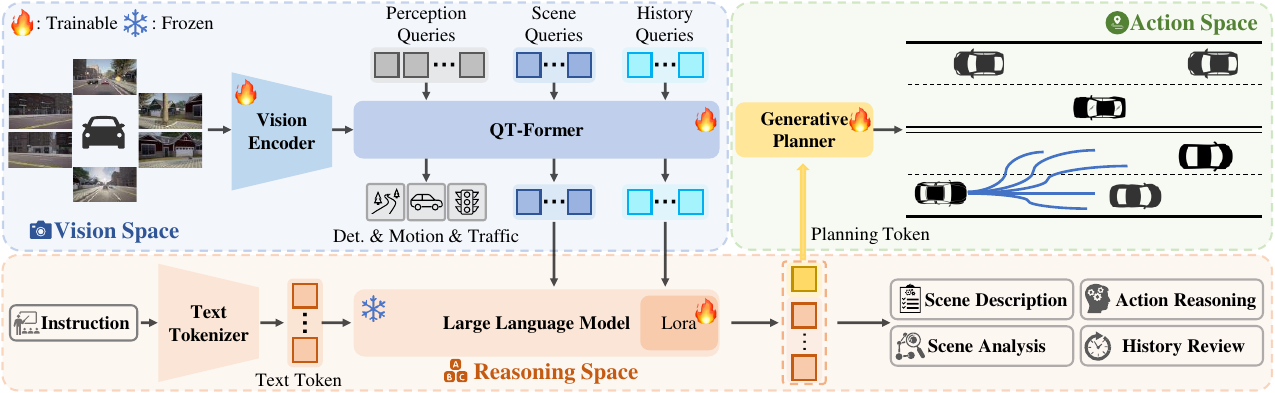}
    % \vspace{-2pt}
    \caption{The pipeline of our ORION, a holistic E2E framework aligning vision-reasoning-action space. It consists of three key components: a QT-Former to extract long-term context and link the vision space of the vision encoder and the reasoning space of LLM; the LLM for performing reasoning tasks and predicting a planning token; and a generative planner that bridges reasoning and action space for generating a multi-modal trajectory conditioned by the planning token.}
    \label{fig: main pipeline}
    % \vspace{-10pt}
\end{figure*}

\subsection{Vision-Language Models (VLMs)}
VLMs~\cite{openai2023gpt, anil2023gemini, liu2023llava, wang2024qwen2vl, chen2024internvl, li2024monkey} introduce visual information to large language models (LLMs)~\cite{touvron2023llama2, lu2024deepseek} through various vision encoders~\cite{radford2021clip,zhai2023sigmoid}, demonstrating powerful visual contextual understanding and reasoning. LLaVA series~\cite{liu2023llava, liu2024llava-next} employ visual instruction tuning to perform image-text alignment. Monkey~\cite{li2024monkey} improves detail comprehension by dividing images. InternVL series~\cite{chen2024internvl, chen2024far} further enhances the vision detail understanding via a dynamic resolution strategy. However, most methods map the numerous visual tokens into language space through MLP, incurring high computational costs. To alleviate this burden, QwenVL~\cite{bai2023qwen} and Flamingo~\cite{alayrac2022flamingo} reduce token redundancy using cross-attention, while Qwen2VL~\cite{wang2024qwen2vl} enhances efficiency with dynamic resolution and multimodal rotary position embedding (M-RoPE) for simultaneously processing diverse modalities. Inspired by these, we introduce QT-Former, which leverages a set of queries and cross-attention operations to extract multi-view image features.

\subsection{VLM for End-to-End Autonomous Driving}
VLMs showcase excellent contextual understanding and comprehensive world knowledge, motivating their application in autonomous driving. Some methods~\cite{wang2024omnidrive,hwang2024emma, xing2025openemma} directly employ VLMs for environment perception and explainable trajectory prediction in text form. For example, Omnidrive~\cite{wang2024omnidrive} adopts StreamPETR~\cite{wang2023streampetr} as Q-Formar3D to compress current scene features and connect the vision-reasoning space and then performs textual trajectory prediction. EMMA~\cite{hwang2024emma}, trained on large-scale data, enables Gemini~\cite{anil2023gemini} to predict discrete textual planning with strong open-loop performance. Other studies~\cite{tian2024drivevlm, jiang2024senna} integrate VLMs with representative E2E models in a fast-slow dual system. DriveVLM~\cite{tian2024drivevlm} leverages VLM to predict the low-frequency trajectory, which will be refined by an E2E model. Senna~\cite{jiang2024senna} further replaces the low-frequency with the meta-action, guiding the VAD~\cite{jiang2023vad} to predict motion. These methods only implement the open-loop evaluation. Although DriveMLM~\cite{wang2023drivemlm} and LMDrive~\cite{shao2024lmdrive} leverage the VLM to implement closed-loop evaluation, they struggle with processing complex scenarios limited by the simple CARLA Town05Long benchmark. In contrast, we propose a holistic E2E framework that employs a generative planner to bridge the reasoning space of VLM and the action space of trajectories, generating precise trajectories with interpretable action decisions in complex real-world driving scenarios of Bench2Drive.

%%%%%%%%%%%%%%%%%%%%%%%%%%%%%%%%%%%%%
%%%%%%%  Method
%%%%%%%%%%%%%%%%%%%%%%%%%%%%%%%%%%%%%%

\section{Method}

In this paper, we propose a h\textbf{O}listic end-to-end autonomous d\textbf{R}iving framework by v\textbf{I}sion-language model instructed acti\textbf{ON} generation, termed ORION. 
The pipeline of our ORION is shown in Fig.~\ref{fig: main pipeline}.  Specifically, given the multi-view images of the current scene, the ORION first encodes the image tokens with a vision encoder. Then, QT-Former (Sec.~\ref{sec: QT Former}) leverages diverse queries to aggregate long-term vision context, compress image tokens, and perceive traffic elements. The LLM (Sec.~\ref{sec: llm}) subsequently combines the compressed scene features and historical vision information with user instructions, performing diverse understanding and reasoning tasks and generating a planning token. Finally, a generative planner (Sec.~\ref{sec: Generate Planner}) bridges the reasoning space of LLM and the action space of trajectories, predicting multi-modal trajectory conditioned by the planning token. ORION effectively aligns the vision-reasoning-action space through these core components, achieving the collaborative optimization of scene understanding and trajectory generation in a unified space.

\begin{figure}[t!]
    \centering
    \includegraphics[width=0.47\textwidth]{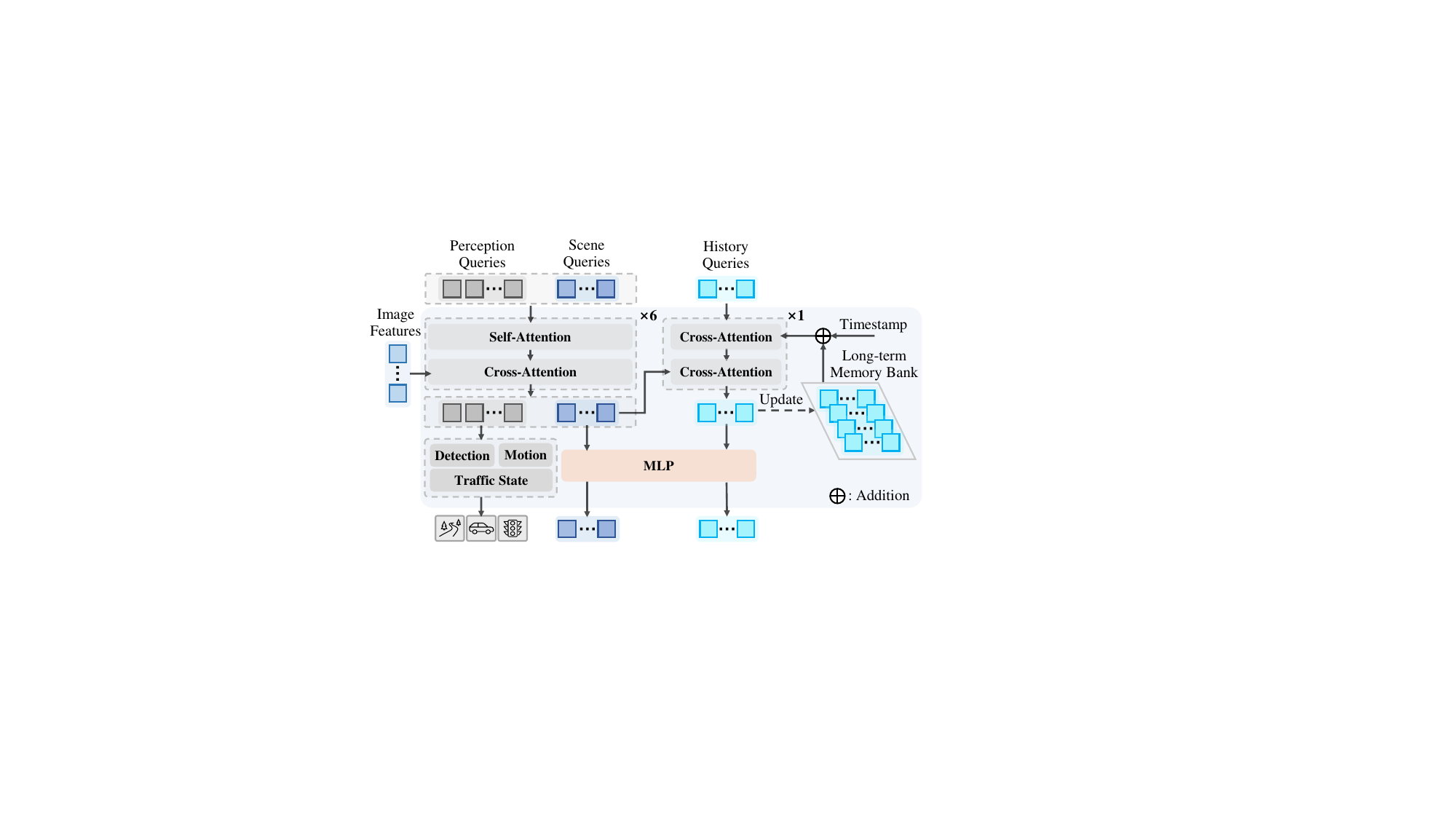}
    % \vspace{-3pt}
    \caption{The detailed architecture of QT-Former. It accepts diverse queries and image features as inputs to detect traffic elements, predict motion, and aggregate long-term vision context.}
    \label{fig: QT Former}
    % \vspace{-9pt}
\end{figure}

\subsection{QT-Former}
\label{sec: QT Former}
To achieve long-term information modeling while compressing and extracting multi-view image features $F_m$ derived from the vision encoder, we introduce QT-Former, a query-based temporal module, as shown in Fig.~\ref{fig: QT Former}. Specifically, following Q-Former3D~\cite{wang2024omnidrive}, we first set up two types of learnable queries, the scene queries $Q_{s} \in \mathbb{R}^{N_s \times C_q}$  and the perception queries $Q_{p} \in \mathbb{R}^{N_p \times C_q}$, where $N_s$ and $N_p$ are the number of scene and perception queries, respectively, and $C_q$ is the channel of queries. ${Q}_s, {Q}_p$ are processed through self-attention (SA) to exchange their information. Then they interact with image features $F_m$ with 3D positional encoding~\cite{liu2022petr} $P_m$ in the cross-attention (CA) module. After that, the perception queries are fed into the multiple auxiliary heads for object detection (\textit{e.g.}, critical objects and lanes), traffic state, and motion prediction of dynamic agents. The scene queries serve as tokens representing the key information of the current scene.

Additionally, we employ a set of history queries $Q_{h} \in \mathbb{R}^{N_h \times C_q}$ and a long-term memory bank $M \in \mathbb{R}^{(N_h\times n) \times C_q}$ to efficiently retrieve and store essential historical information (\textit{e.g.}, previous road conditions and ego status), where $N_h$ is the number of history queries and $n$ is the maximum history frame length. We utilize the $Q_{h}$ to extract the former frame queries in $M$ with relative timestamp embedding $P_t$ through a CA block. Then $Q_h$ interacts with current scene features $Q_s$ in another CA block, enabling the extraction of relevant details about the current scenario. This process can be formulated as:
\begin{equation}
\begin{aligned}
    Q_h &= \text{CA}(Q_h, M + P_t, M + P_t), \\
    \hat{Q}_h &= \text{CA}(Q_h, Q_s, Q_s),
\end{aligned}
\end{equation}
where $P_t$ denotes the relative timestamp embedding.  

Subsequently, the updated history queries $\hat{Q}_h$ are stored in the memory bank $M$ following the First-In-First-Out (FIFO) replacement policy, formulated as:
\begin{equation}
\begin{aligned}
   % M = [Q^{t-n}_{h},\cdot \cdot \cdot, Q^{t-1}_{s} + T^{t-1}, Q^{t}_{s} + T^{t}],
   M = [\hat{Q}^{t-n}_{h}, \cdot \cdot \cdot, \hat{Q}^{t-1}_{h}, \hat{Q}^{t}_{h}] ,
\end{aligned}
\end{equation}
where $t$ is the current frame time.

Although some methods~\cite{wang2023streampetr,song2024moviechat} also leverage the memory bank to store preceding information, they typically 
only store the compressed historical information without guiding for extracting the current scene information. Instead, we initialize a few numbers of the history queries to further extract the current scene features that are most closely related to historical information, enhancing the long-term memory ability of the model.

Finally, we utilize a two-layer MLP to convert the updated history queries $\hat{Q}_h$ and current scene features $Q_s$ to history tokens $x_h$ and scene tokens $x_s$ in the reasoning space of LLM.

\subsection{Large Language Model}
\label{sec: llm}
The LLM is pivotal in our framework because the high-quality reasoning of the current driving scenario is necessary to instruct the generative planner to generate a reasonable trajectory in action space.

As shown in Fig.~\ref{fig: main pipeline}, the user instruction is first encoded into language tokens $x_q \in \mathbb{R}^{L \times C}$ by the text tokenizer, where $L$ is the token length and $C$ is the dimension of LLM. Then, the scene tokens $x_s$ and history tokens $x_h$ are combined with the language tokens $x_q$ and fed into LLM. 

Leveraging the abundant world knowledge and outstanding reasoning ability of LLM, ORION performs various text-based understanding and reasoning tasks in the driving scenario, including scene description, history information review, scene analysis, and action reasoning. Meanwhile, we design a planning QA template with a special planning token $s$ for LLM as the final QA to accumulate the understanding and reasoning context of the entire driving scenario to the $s$, formally written as:
\begin{equation}
    s \sim p(s|x_s, x_h, x_q, x_a),
\end{equation}
where $x_a$ denotes the generation answer of LLM. The embedding of the planning token $s$ will serve as a condition to control the trajectory generation.

However, there is still a lack of high-quality VQA annotations within closed-loop simulation environments to train LLMs for comprehensively understanding driving scenarios. Thus, we extend the Bench2Drive dataset via a fully automatic VQA annotation pipeline powered by Qwen2-VL~\cite{wang2024qwen2vl} and propose our VQA dataset, Chat-B2D, expecting to further promote the research of VLM on closed-loop simulation. We provide detailed information on Chat-B2D and its annotation pipeline in the Appendix.

\subsection{Generative Planner}
\label{sec: Generate Planner}
Generative models~\cite{kingma2013auto,ronneberger2015u,goodfellow2020generative} can effectively capture intrinsic features within data by learning the distribution of the data. Recent researches~\cite{rombach2022high, betker2023improving, liu2024sora} have demonstrated semantic correlations between latent spaces of different modalities of data, where adjusting the distribution parameters of one modality space enables precise control over the generation process of another modality space.

Inspired by the generative domain, we introduce a generative planner to bridge the gap between the reasoning and action space. Specifically, we formulate the current trajectory $a$ in action space as a conditional probability distribution $p(a|s)$, where $s$ is the planning token. To construct $p(a|s)$, there are many excellent methods in the generation field (\textit{e.g.}, variational autoencoders (VAE)~\cite{kingma2013auto} and diffusion model~\cite{ronneberger2015u}). 

As there are essential differences in the distribution between the reasoning space of VLM and the action space of trajectory, we use the VAE~\cite{kingma2013auto} model to align them in the Gaussian distribution. We employ two-layer MLPs to project both the state $s$ and the ground-truth trajectory $t$ into Gaussian variables $z$ in the latent space, denoted as:
\begin{equation}
    p(z_s|s) \sim N(\mathbf{\mu}_s,\mathbf{\sigma}_s^2),  
    p(z_t|t) \sim N(\mathbf{\mu}_t,\mathbf{\sigma}_t^2),
\end{equation}
where $N(\mathbf{\mu_{}},\mathbf{\sigma}^2)$ denotes a Gaussian distribution with a mean of $\mathbf{\mu},$ and standard deviation of $\mathbf{\sigma}$. We then use Kullback-Leibler divergence loss to enforce distribution matching, represented as:
\begin{equation}
    \mathcal{L}_{vae} = D_{KL}(p(\mathbf{z}|\mathbf{s}), p(\mathbf{z}|\mathbf{t})).
\end{equation}

Finally, we use the GRU decoder in GenAD~\cite{zheng2024genad} to decode the trajectory from the latent space $z$. 
Significantly, the functions of VAE in this paper are not the same as VAE of GenAD. We only use a single token encoded in the reasoning space from the perspective of the ego vehicle as input, aiming to bridge the gap between reasoning space and action space. In contrast, the latter leverages features of all agents encoded in the BEV space as input, designed to learn specific patterns of the highly structured trajectories of both the ego vehicle and other agents.

Additionally, we also attempt to replace the VAE with alternative generative models, such as the diffusion model for trajectory generation. Benefiting from the proposed method that bridges the gap between the reasoning and action space through distribution learning in latent space, our framework still demonstrates superior performance compared to other methods (detailed in Sec.~\ref{sec: Ablation Study}). 

\subsection{Training Objectives}
For the detection task of the proposed QT-Former, the detection loss is defined as $\mathcal{L}_{det} = \mathcal{L}_{cls} + \mathcal{L}_{reg}$, where $\mathcal{L}_{cls}$ is focal loss~\cite{lin2017focal} and $\mathcal{L}_{reg}$ is L1 loss. For the traffic state and motion prediction, the losses are defined as $\mathcal{L}_{tra}$ and $\mathcal{L}_{m} = \mathcal{L}_{mcls} + \mathcal{L}_{mreg}$, respectively, where $\mathcal{L}_{tra}$ and $\mathcal{L}_{mcls}$ are focal loss, and $ \mathcal{L}_{mreg}$ is L1 loss. The total loss of QT-Former is:
\begin{equation}
    \mathcal{L}_{qt} = \mathcal{L}_{det} + \mathcal{L}_{tra} + \mathcal{L}_{m}.
\end{equation}

For the LLM, we leverage the auto-regressive cross-entropy loss $\mathcal{L}_{ce}$.
For the generative planner in our framework, $\mathcal{L}_{vae}$ is the Kullback-Leibler divergence loss used to align the reasoning space and action space. Following VAD~\cite{jiang2023vad}, we adopt the collision loss $\mathcal{L}_{col}$, boundary loss  $\mathcal{L}_{bd}$, and MSE loss $\mathcal{L}_{mse}$ for the planning prediction. The total loss of the generative planner is:
\begin{equation}
    \mathcal{L}_{gp} = \mathcal{L}_{vae} + \mathcal{L}_{mse} + \mathcal{L}_{col} + \mathcal{L}_{bd}.
\end{equation}

In summary, the total loss of the proposed ORION is:
\begin{equation}
    \mathcal{L} = \mathcal{L}_{qt} + \mathcal{L}_{ce} + \mathcal{L}_{gp}.
\end{equation}

%%%%%%%%%%%%%%%%%%%%%%%%%%%%%%%%%%%%%
%%%%%%%  Experiment
%%%%%%%%%%%%%%%%%%%%%%%%%%%%%%%%%%%%%%

\section{Experiments}
\label{sec: Experiments}

\subsection{Dataset and Evaluation Metrics}
\noindent\textbf{Dataset.}
We train and evaluate ORION on the Bench2drive dataset~\cite{jia2024bench2drive}, a closed-loop evaluation protocol under CARLA V2~\cite{dosovitskiy2017carla} for E2E autonomous driving.
It provides an official training set where we use the base set (1000 clips) for fair comparison with all the other baselines, which is divided into 950 clips for training and 50 clips for open-loop validation. Each clip captures approximately 150 meters of continuous driving within a specific traffic scene. For closed-loop evaluation, we evaluate the proposed method on the official set of 220 short routes designed by Bench2drive, spanning 44 interactive scenarios with 5 routes per scenario. Additionally, we compare our method with other SOTA baselines on nuScenes~\cite{caesar2020nuscenes} open-loop evaluation, which will be provided in the Appendix.

\noindent\textbf{Evaluation Metrics.}
Bench2drive includes five metrics for closed-loop evaluation: Driving Score (DS), Success Rate (SR), Efficiency, Comfortness, and Multi-Ability. The Success Rate quantifies the proportion of routes successfully completed within the allotted time. The Driving Score follows CARLA~\cite{dosovitskiy2017carla}, incorporating both route completion status and violation penalties, where infractions reduce the score via discount factors. Efficiency and Comfortness are used to measure the speed performance and comfort of the autonomous driving system during the driving process, respectively. Multi-Ability measures 5 advanced skills independently for urban driving. For open-loop evaluation, we use the L2 distance error and the collision rate. Additionally, we use CIDEr~\cite{vedantam2015cider}, BLEU~\cite{papineni2002bleu}, and ROUGE-L~\cite{lin2004rouge} to evaluate the performance of ORION on VQA tasks.

\begin{table*}
\centering
\caption{Closed-loop and Open-loop Results of E2E-AD Methods in Bench2Drive under base set. C/L refers to camera/LiDAR. Avg. L2 is averaged over the predictions in 2 seconds under 2Hz, similar to UniAD. * denote expert feature distillation. NC: navigation command, TP: target point, DS: Driving Score, SR: Success Rate.
\label{tab: main result}}
% \resizebox{1.0\textwidth}{!}{\begin{tabular}{l| c | c c c c}
\footnotesize
\setlength\tabcolsep{1.6 mm} %列
\begin{tabular}{l c c c  >{\columncolor{gray!10}}c >{\columncolor{gray!10}}c c c   c}
\toprule
\multirow{2.2}{*}{Method} & \multirow{2.2}{*}{Reference} &\multirow{2.2}{*}{Condition} &\multirow{2.2}{*}{Modality}  & \multicolumn{4}{c}{Closed-loop Metric}  & \color{gray}Open-loop Metric  \\  
\cmidrule(lr){5-8}  \cmidrule(lr){9-9} & & &  &  DS$\uparrow$   & SR(\%)$\uparrow$ & Efficiency$\uparrow$  & Comfortness$\uparrow$ &  \color{gray}Avg. L2 $\downarrow$\\ \midrule

TCP*~\cite{wu2022trajectoryguided}& NeurIPS 22 & TP & C   &  40.70     & 15.00  & 54.26 & 47.80  & \color{gray}1.70   \\
TCP-ctrl* &NeurIPS 22  & TP    & C   &  30.47    & 7.27  & 55.97 & 51.51 & \color{gray}-  \\
TCP-traj* & NeurIPS 22& TP & C   &  59.90     & 30.00 & 76.54 & 18.08    & \color{gray}1.70   \\
TCP-traj w/o distillation & NeurIPS 22  & TP   & C    &  49.30     & 20.45  & 78.78 & 22.96  & \color{gray}1.96   \\
ThinkTwice*~\cite{jia2023thinktwice}  & CVPR 23 & TP & C   & 62.44     & 31.23  & 69.33 & 16.22   & \color{gray}0.95 \\
DriveAdapter*~\cite{jia2023driveadapter} &ICCV 23 & TP & C\&L   &  64.22   & 33.08 & 70.22 & 16.01    & \color{gray}1.01 \\  

\midrule

AD-MLP~\cite{zhai2023rethinking} & arXiv 23 &  NC &C    & 18.05     &  0.00  & 48.45 &   22.63  & \color{gray}3.64\\
UniAD-Tiny~\cite{hu2023planning} & CVPR 23 & NC & C   & 40.73    & 13.18 & 123.92 & 47.04  &  \color{gray}0.80  \\
UniAD-Base~\cite{hu2023planning} & CVPR 23 & NC  & C  & 45.81     & 16.36 & 129.21 & 43.58  & \color{gray}0.73  \\

VAD~\cite{jiang2023vad}& ICCV 23 & NC & C  &42.35     & 15.00 & 157.94 & 46.01  &  \color{gray}0.91\\ 

GenAD~\cite{zheng2024genad} & ECCV 24 & NC &C & 44.81 & 15.90 &  -& - &\color{gray}-\\

MomAD\cite{song2025don} & CVPR25 & NC & C & 44.54	&16.71&	170.21&	48.63& \color{gray}0.87\\
DriveTransformer-Large~\cite{jiadrivetransformer} & ICLR 25 & NC   & C & 63.46     & 35.01 & 100.64 & 20.78  &  \color{gray}\textbf{0.62} \\ 

\midrule
\rowcolor[RGB]{230, 242, 255}ORION(\textbf{Ours})  & - & NC & C & \textbf{77.74}\dplus{+14.28}  & \textbf{54.62}\dplus{+19.61} & 151.48& 17.38 & \color{gray}0.68   \\

\bottomrule
\end{tabular}
\end{table*}

\begin{table*}[htbp]
\centering
\caption{Multi-Ability Results of E2E-AD Methods under base set.  * denote expert feature distillation. C/L refers to camera/LiDAR. NC: navigation command, TP: target point.
\label{tab: multi-ability}}

% \resizebox{1.0\textwidth}{!}
\footnotesize
\setlength\tabcolsep{1.2mm} %列
{\begin{tabular}{l c c c  c c c c c >{\columncolor{gray!10}}c}
\toprule
\multirow{2.3}{*}{Method} & \multirow{2.3}{*}{Reference} & \multirow{2.3}{*}{Condition}  & \multirow{2.3}{*}{Modality} & \multicolumn{5}{c}{Ability (\%) $\uparrow$} \\ 
\cmidrule{5-10} & & & & \multicolumn{1}{c}{Merging} & \multicolumn{1}{c}{Overtaking} & \multicolumn{1}{c}{Emergency Brake} & \multicolumn{1}{c}{Give Way} & Traffic Sign & Mean \\ \midrule
TCP*~\cite{wu2022trajectoryguided}  &NeurIPS 22 &TP &C  & 16.18        & 20.00           & 20.00        &  10.00         & 6.99     & 14.63         \\
TCP-ctrl*  & NeurIPS 22 &TP &C & 10.29        & 4.44           & 10.00        &  10.00          & 6.45     & 8.23         \\
TCP-traj*   &NeurIPS 22 &TP &C & 8.89       & 24.29           & 51.67       &  40.00        & 46.28    & 34.22        \\ 
TCP-traj w/o distillation  & NeurIPS 22 &TP &C & 17.14       & 6.67           & 40.00       &  \textbf{50.00}    & 28.72    & 28.51        \\ 
ThinkTwice*~\cite{jia2023thinktwice}    & CVPR 23  &TP &C & 27.38       &  18.42          & 35.82       &  \textbf{50.00}        & 54.23    & 37.17       \\ 
DriveAdapter*~\cite{jia2023driveadapter}  &  ICCV 23 &TP & C\&L & \textbf{28.82}       &26.38           & 48.76      &  \textbf{50.00}         & 56.43   & 42.08        \\
\midrule

AD-MLP~\cite{zhai2023rethinking} &  arXiv 23 & NC & C & 0.00        & 0.00           & 0.00        & 0.00         &  4.35    & 0.87         \\
UniAD-Tiny~\cite{hu2023planning} & CVPR 23 & NC & C & 8.89        & 9.33           & 20.00       & 20.00        & 15.43    & 14.73           \\ 
UniAD-Base~\cite{hu2023planning}  & CVPR 23 & NC & C  & 14.10       & 17.78          & 21.67       &  10.00       & 14.21    & 15.55       \\ 
VAD~\cite{jiang2023vad} &   ICCV 23  & NC & C  & 8.11       & 24.44         & 18.64       &  20.00       & 19.15    & 18.07       \\ 
DriveTransformer-Large~\cite{jiadrivetransformer}& ICLR 25 & NC & C  & 17.57 & 35.00 & 48.36 & 40.00 &52.10 &38.60 \\

\midrule

\rowcolor[RGB]{230, 242, 255}ORION (\textbf{Ours}) &- & NC & C  & 25.00 & \textbf{71.11} &\textbf{78.33}  &30.00 &\textbf{69.15} &\textbf{54.72}\dplus{+16.12}\\

\bottomrule
\end{tabular}}
% \vspace{-3mm}
\end{table*}

\subsection{Implementation Details}
\noindent\textbf{Model Setting.}
Consistent with classic E2E baselines~\cite{hu2023planning, jiang2023vad, zheng2024genad} on Bench2Drive, ORION is a fully HD map-free method that only uses the Navigation Command (NC) as an input condition for the trajectory predictions rather than locations of lane center (\textit{i.e.}, target point, TP). ORION is an anchor-free method that outputs 6 mode trajectory predictions corresponding to the 6 NC defined in Bench2Drive. 

\noindent\textbf{Training Process.} All experiments are conducted on 32 NVIDIA A800 GPUs with 80 GB of memory. Following Omnidrive ~\cite{wang2024omnidrive}, we adopt EVA-02-L~\cite{fang2024eva} as the vision encoder. Vicuna v1.5~\cite{zheng2023judging} is employed in ORION and fine-tuned using LoRA~\cite{hu2021lora}, with the rank dimension and alpha set to 16. The default number of scene, perception, and historical queries is 512, 600, and 16, respectively. We set the Memory Bank's stored frame number $n$ to 16. During training, data augmentations are applied to input images, which are first resized to a resolution of $640 \times 640$. More training details are provided in the Appendix.

\subsection{Main Results}

As reported in Tab.~\ref{tab: main result}, the performance of ORION significantly exceeds all E2E methods on Bench2Drive, even the method with expert feature distillation. Specifically, ORION surpasses the latest SOTA method  DriveTransformer~\cite{jiadrivetransformer} by +14.28 DS and +19.61\% SR. It also achieves improvements of +13.52 DS and +21.54\% SR over DriveAdapter~\cite{jia2023driveadapter}, even if DriveAdapter distills the expert feature from Think2Drive~\cite{li2024think2drive} and leverages two modalities (\textit{i.e.}, camera and LiDAR) inputs. The above promising results effectively demonstrate the superiority of our ORION.

Additionally, the Multi-Ability results are also illustrated in Tab.~\ref{tab: multi-ability}. ORION achieves +16.12\% and +12.64\% performance improvements compared with DriveTransformer~\cite{jiadrivetransformer} and DriveAdapter~\cite{jia2023driveadapter} in the mean ability, respectively. Specifically, our model demonstrates outstanding performance in some scenarios, such as Overtaking (71.11\%), Emergency Brake (78.33\%), and Traffic Sign (69.15\%), which shows that our model benefits from the powerful reasoning capability of VLM to understand the causal interaction between the ego vehicle, dynamic elements, and static elements (Traffic Signs) in driving scenarios. On the other hand, our model falls behind DriveAdapter in Merging and Give Way, which shows that ORION is not good at making lane-changing decisions. The phenomenon may be caused by the more diverse decision-making timing for lane-changing, making the model encounter difficulties in capturing the correct causal relationship~\cite{jia2023driveadapter}.

\begin{figure*}[htbp!]
    \centering
    \includegraphics[width=\textwidth]{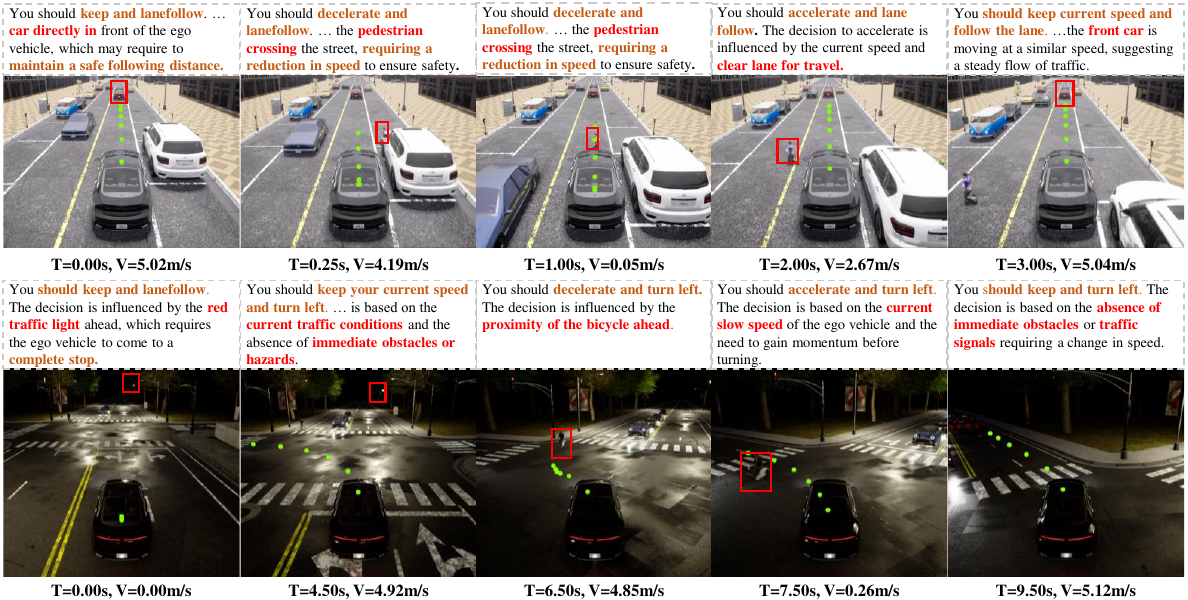}
    \caption{Qualitative results of ORION on the Bench2Drive closed-loop evaluation set. The \textcolor{brown}{brown}, \textcolor{red}{red}, and \textcolor{green}{green} refer to the action decision, the objects that influence driving decisions, and the prediction trajectory, respectively.}
    \label{fig: qualitative results}
\end{figure*}

\begin{figure}[t]
    \centering
    \includegraphics[width=0.47\textwidth]{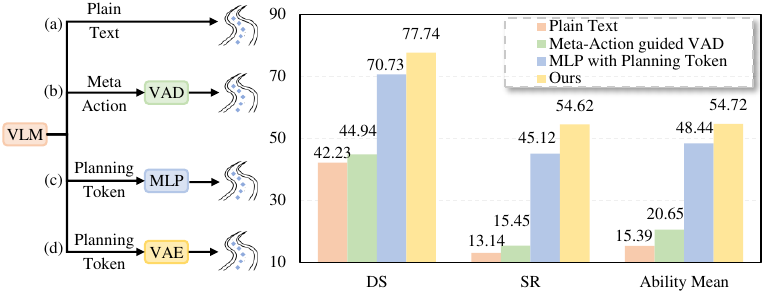}
    \caption{Advantages of the vision-language instructed action generation. DS and SR denote Driving Score and Success Rate separately. VAD~\cite{jiang2023vad} is a classic E2E model.} 
    \label{fig: alignment method}
    % \vspace{-10pt}
\end{figure}

\subsection{Qualitative Results}
The qualitative results of ORION in two canonical closed-loop evaluation scenarios of Bench2Drive are shown in Fig.~\ref{fig: qualitative results}. It shows both the driving action reasoning and trajectory prediction outputted by our model, as well as the corresponding ego-vehicle states. We observe that ORION can capture the correct causal relationship in the scenario and make correct driving decisions, then predict the planning trajectory following the reasoning instruction, demonstrating the surprising interpretability of our method. More qualitative results can be found in the Appendix.

\subsection{Ablation Study}
\label{sec: Ablation Study}

\noindent\textbf{Advantages of the vision-language instructed action generation.} To validate the effectiveness of the planning generation paradigm proposed in this paper, extensive experiments are conducted to compare our paradigm with canonical trajectory prediction paradigms of VLM-based E2E autonomous driving methods, including (a) plain text outputs~\cite{wang2024omnidrive,hwang2024emma}, (b) dual-system paradigm which classic E2E methods(\textit{e.g.}, VAD~\cite{jiang2023vad}) output trajectory guided by elaborated design VLM interface (\textit{e.g.}, meta-action)~\cite{jiang2024senna}, and (c) special token decode outputs by MLP~\cite{renz2024carllava}, as shown in the left part of Fig.~\ref{fig: alignment method}. To ensure the fairness of the ablations, experiments of different paradigms use the same sensor inputs, vision encoder, QT-former, and VLM as our ORION and are trained by the same strategy. Only the output formats of VLMs are adjusted according to the requirements of different paradigms. 

The results are illustrated in the right part of Fig.~\ref{fig: alignment method}. The plain text paradigm performs the worst (42.23 DS, 13.14\% SR, and 15.39\% mean ability), indicating the limitations of plain text output in closed-loop driving scenarios, potentially due to its inadequate numerical reasoning capabilities~\cite{frieder2024mathematical, peng2021mathbert}. Compared with the plain text paradigm, the dual-system paradigm only obtains a slight performance improvement. Note that the reproduced results of the dual-system paradigm are very close to the official results of VAD in Tab.~\ref{tab: main result}. This result may indicate that the performance of the dual-system paradigm may be bottlenecked by the insufficient capabilities of classic E2E methods. Although the effectiveness of the MLP decoder paradigm has been validated in CarLLaVA~\cite{renz2024carllava}, our paradigm still shows a performance gain of +7.01 DS, +9.5\% SR, and +6.28\% mean ability. The result may be caused by the fact that the MLP is the simplest way to align features between different spaces, which is consistent with the viewpoint presented in this paper. Additionally, the MLP decoder struggles with handling multi-modal trajectory~\cite{jaeger2023hidden,chen2024vadv2}, making it still significantly lag behind ORION in closed-loop evaluation.

\begin{table}[tp!]
\centering
\caption{Ablation on diverse generative planner. DS and SR denote Driving Score and Success Rate separately.}
\label{tab: generative planner}
% \resizebox{1.0\textwidth}{!}{\begin{tabular}{l| c | c c c c}
\footnotesize
\setlength\tabcolsep{1.0mm} %列
\begin{tabular}{ c   c c  c c  c}
\toprule
\multirow{2.6}{*}{\shortstack{Generative\\Planner}}  & \multicolumn{2}{c}{Closed-loop} & \multicolumn{2}{c}{\color{gray} Open-loop} & Ability\\ 
\cmidrule(lr){2-3} \cmidrule(lr){4-5} \cmidrule(lr){6-6} & DS$\uparrow$ & SR(\%)$\uparrow$  &\color{gray} Avg. L2 (m) $\downarrow$ & \color{gray} Avg. col (\%)$\downarrow$ & Avg.\\ 
\midrule
Diffusion &  71.97   &    46.54  & \color{gray}0.73 & \color{gray}0.96 & 46.68\\
VAE (Ours)   &  77.74   & 54.62   & \color{gray}0.68  & \color{gray}0.47 &54.72\\
\bottomrule
\end{tabular}
\end{table}

\begin{table}[tp!]
\centering
\caption{Ablation on QT-Former designs in different frameworks. DS and SR denote Driving Score and Success Rate separately. Traffic state means using explicit traffic state supervision. T: Plain Text, G: Instructed Generator}
\label{tab: model ablation}
% \resizebox{1.0\textwidth}{!}{\begin{tabular}{l| c | c c c c} 
\footnotesize
\setlength\tabcolsep{1.6mm} %列
\begin{tabular}{c  c c c c c   c c }
\toprule
\multirow{2.3}{*}{ID} & \multirow{2.6}{*}{\shortstack{Traffic\\State}} & \multirow{2.6}{*}{\shortstack{Motion\\Pred.}}  & \multirow{2.6}{*}{\shortstack{Memory \\ Bank}}  
& \multicolumn{2}{c}{Output type}  & \multicolumn{2}{c}{Closed-loop} \\ 
\cmidrule(lr){5-6} \cmidrule(lr){7-8} & & & & T & G & DS $\uparrow$  & SR(\%) $\uparrow$  \\ 
\midrule
1 &   &   &     & & \checkmark &   56.33   & 26.05  \\
2 & \checkmark  &    &  &  & \checkmark &  74.65    & 49.31 \\
3 & \checkmark  & \checkmark  &  &  &\checkmark &  74.07   &  49.77   \\
4 & \checkmark   &\checkmark    &\checkmark  & & \checkmark  &  77.74  & 54.62  \\

5 &  &   &  &  \checkmark &  &  25.45  & 10.38  \\
6 & \checkmark   &  \checkmark & \checkmark & \checkmark  &  &  42.23 & 13.14  \\
\bottomrule
\end{tabular}
\end{table}

\noindent\textbf{Analysis on different generative planners.} We then investigate the effect of employing different generative planners to bridge the reasoning-action space. Specifically, we implement the diffusion model by simply replacing the VAE, which uses K-means trajectory anchors as prior information and outputs 20 mode trajectory predictions. The results are listed in Tab.~\ref{tab: generative planner}. Note that the VAE-based trajectory generator demonstrates a significant performance improvement over the diffusion-based. We argue the main reasons are as follows: 1) Compared with the conditional denoising process of diffusion, the latent space of VAE more directly and effectively aligns the reasoning information of VLM to the multi-modal action space. 2) The training process of VAE is inherently more stable, facilitating better alignment between the reasoning and action spaces. Surprisingly, even using diffusion, ORION still surpasses the DriveTransformer by +8.51 DS, +11.53\% SR, and +8.08\% mean ability. This impressive result emphasizes the effectiveness and flexibility of our framework.
% in Tab.~\ref{tab: main result} and Tab.~\ref{tab: multi-ability}

\noindent\textbf{Effectiveness of QT-Former designs.} Tab.~\ref{tab: model ablation} shows the detailed ablations of each design in the introduced QT-Former. By leveraging explicit traffic state supervision (ID-2),  ORION achieves 74.65 DS and 49.31\% SR, which already outperforms DriveAdapter~\cite{jia2023driveadapter} and DriveTransformer~\cite{jiadrivetransformer} by a large margin and makes an improvement of +18.32 and +23.26\% compared with the baseline (ID-1). This is because a better understanding of traffic signals helps ORION directly reduce infractions in closed-loop evaluation. It is worth noting that due to the causal confusion~\cite{jia2023driveadapter}, it's not trivial for previous methods to fully understand the corresponding causal relationships by simply introducing traffic state supervision, especially when encountering mixed expert behaviors before traffic signs~\cite{jia2023driveadapter,jiadrivetransformer,jia2023thinktwice,wu2022trajectoryguided}. This result also proves that ORION can better utilize the reasoning ability of VLM to capture the causal relationship between scene information and driving behavior by aligning reasoning space and action space. This conclusion can also be verified by the results in Tab.~\ref{tab: multi-ability}, where ORION shows a significant advantage in traffic sign ability (+17.05\%) compared to previous E2E methods~\cite{jiadrivetransformer}.

\begin{table}[tp!]
\centering
\caption{Ablation of history queries number. DS and SR denote Driving Score and Success Rate separately.}
\label{tab: historical token number ablation}
% \resizebox{1.0\textwidth}{!}{\begin{tabular}{l| c | c c c c}
\footnotesize
\setlength\tabcolsep{1.5mm} %列
\begin{tabular}{c  c c   c  c}
\toprule
\multirow{2.2}{*}{Query Num. $N_h$}  & \multicolumn{2}{c}{Closed-loop}  & \multicolumn{2}{c}{\color{gray}Open-loop} \\ 
\cmidrule(lr){2-3} \cmidrule(lr){4-5}  & DS $\uparrow$  & SR(\%) $\uparrow$ & \color{gray}Avg. L2 (m) $\downarrow$ & \color{gray}Avg. col (\%)$\downarrow$ \\ 
\midrule
% 4    &&   & -   &-    &-  & \color{gray}-\\
0     & 65.10 &38.83    &\color{gray}0.67 &\color{gray}0.61 \\
8     & 68.09& 39.09   &\color{gray}0.66 &\color{gray}0.62 \\
16    & 74.10  & 44.66    &\color{gray} 0.68 &\color{gray} 0.55 \\ 
32    & 62.46& 37.73    &\color{gray}0.65 & \color{gray}0.73\\

\bottomrule
\end{tabular}
\end{table}

Then, we combine the motion prediction module in the QT-Former's perception head, which gains a slight improvement of +0.4\% SR and further reduces the collision rate. The slight degradation on DS may be caused by the trade-off between DS and SR in the CARLA benchmark protocol~\cite{zimmerlin2024hidden}. Involving a memory bank into QT-Former and supervised by QA pairs about historical information leads to an increase of +3.67 DS and +4.85\% SR and boosts the final performance to 77.74 DS and 54.62\% SR, which demonstrates our model can benefit from the long-temporal memory of vision tokens.

We also apply QT-former to the plain text output type (ID-6). By leveraging it, we improve the model's performance by +16.78 DS and +2.78\% SR over the baseline (ID-5). Meanwhile, with the same QT-former designs, our ORION achieves further improvements of +35.51 DS and +41.48\% SR compared with the plain text output mode, demonstrating the effectiveness of our framework.

\noindent\textbf{Influence of the number of history queries.} We conduct ablation experiments to further study the influence of different numbers of history queries. Here, to accelerate the training process, we only train the model using the planning trajectory and history QA pairs without other auxiliary VQA tasks. The results are detailed in Tab.~\ref{tab: historical token number ablation}. Increasing the history query number $N_h$ from 0 to 8 brings a significant performance boost of around 2.99 DS and 0.26\% SR. Further increasing $N_h$ from 8 to 16 leads to the sweet point that achieves the best performance of 74.10 DS and 44.66\% SR. However, enlarging $N_h$ from 16 to 32 shows a significant performance degradation. We argue that introducing more history queries hinders the VLM from capturing the current frame features, which are more essential than historical information in the driving scene.

\begin{table}[tp!]
\centering
\caption{Effectiveness of auxiliary VQA task training. DS and SR denote Driving Score and Success Rate separately. C/B/R refers to CIDEr/BLEU/ROUGE-L. FT: Fine Tuning}
\label{tab: training alignment}
% \resizebox{1.0\textwidth}{!}{\begin{tabular}{l| c | c c c c}
\footnotesize
\setlength\tabcolsep{0.7mm} %列
\begin{tabular}{c  c c   c c  c c c   c }
\toprule
\multirow{2.3}{*}{ID} & \multirow{2.6}{*}{\shortstack{VQA\\FT}} & \multirow{2.6}{*}{\shortstack{Planning\\FT}}  & \multicolumn{2}{c}{Closed-loop}  &\multicolumn{3}{c}{Language}  & \color{gray}Open-loop\\ 
\cmidrule(lr){4-5} \cmidrule(lr){6-8} \cmidrule(lr){9-9} &  &  & DS $\uparrow$  & SR(\%) $\uparrow$ &C$\uparrow$ &B$\uparrow$ &R$\uparrow$ & \color{gray}Avg. L2 (m) $\downarrow$  \\ 
\midrule
1 & \checkmark  &        &- & -  &65.65  &50.82  &77.65  & - \\
2 &  & \checkmark        &74.10 & 44.66  & -   &-    &- &  \color{gray}0.68 \\
3 &\checkmark  & \checkmark   &  77.74 & 54.62   & 65.77  & 52.49  & 77.58  & \color{gray}0.68 \\ 

\bottomrule
\end{tabular}
\end{table}

\noindent\textbf{Influence between VQA task training and planning task training.} As shown in Tab.~\ref{tab: training alignment}. The model cannot obtain both reasoning and planning capabilities with single-task training. Surprisingly, when we perform two tasks simultaneously during training, ORION achieves better performance in both planning and language metrics compared to single-task training. Specifically, the multi-task training leads to improvements of +3.64 DS and +9.66\% SR in the planning task, as well as a performance gain of +0.12 CIDEr, +1.67 BLEU and competitive performance of ROUGE-L in the VQA tasks. Furthermore, the results also validate the high quality and validity of the Chat-B2D dataset produced by our auto-pipeline.

%%%%%%%%%%%%%%%%%%%%%%%%%%%%%%%%%%%%%
%%%%%%%  Conclusion
%%%%%%%%%%%%%%%%%%%%%%%%%%%%%%%%%%%%%%

\section{Conclusion}
\label{sec: conclusion}
In this paper, we mainly focus on the challenges faced by VLM methods for end-to-end autonomous driving in aligning the reasoning space of VLM with the pure numerical action space used for planning. This dilemma makes it not trivial for existing methods to simultaneously analyze the driving scenario and output high-quality multimodal prediction trajectories. To address this problem, we propose ORION,  a holistic end-to-end autonomous driving framework by vision-language instructed action generation. By leveraging a generative planner and incorporating long-term visual context, we effectively bridge the vision-reasoning-action space.  Extensive experiments validate the flexibility and superiority of our proposed framework, where ORION demonstrates significant improvements in closed-loop planning evaluation, surpassing SOTA methods.

\noindent\textbf{Limitation.} Although ORION performs well in the closed-loop simulation environment on Bench2Drive~\cite{jia2024bench2drive}, it is limited by the high computational complexity of the scalable VLM in real-time driving scenarios. In the future, we would like to reduce the complexity of ORION through techniques such as model compression and pruning, thereby enabling the model to achieve real-time autonomous driving.

{
    \small
    \bibliographystyle{ieeenat_fullname}
    \bibliography{main}
    
}
\clearpage
\setcounter{page}{1}
\setcounter{figure}{0}
\setcounter{table}{0}
\renewcommand\thefigure{A\arabic{figure}} 
\renewcommand\thetable{A\arabic{table}}
\maketitlesupplementary
\appendix

\begin{table*}[htbp!]
% \vspace{-2mm}
\centering
\caption{
Comparison of the Open-loop planning in nuScene. \dag: The ego status and planning trajectory are both processed by LLM in textual modality. ${}\ddagger$: The high-level command is not used during the training and testing phases.}
\label{tab: nuscene results}
\footnotesize
\setlength\tabcolsep{3.3 mm} %列
\begin{tabular}{l c cc  cccc  cccc}
\toprule
\multirow{2.2}{*}{Method} &
\multirow{2.2}{*}{VLM-Based} &
\multicolumn{2}{c}{Ego Status} &
\multicolumn{4}{c}{L2 (m) $\downarrow$} & 
\multicolumn{4}{c}{Collision (\%) $\downarrow$}\\
 \cmidrule(lr){3-4} \cmidrule(lr){5-8} \cmidrule(lr){9-12} & &BEV &Planner& 1s & 2s & 3s &Avg. & 1s & 2s & 3s& Avg.\\
\midrule
ST-P3 &- &- &- & 1.33 & 2.11 & 2.90 & 2.11 & 0.23 & 0.62 & 1.27 & 0.71 \\
UniAD~\cite{hu2023planning}&- &-  & - & 0.48 & 0.96 & 1.65 & 1.03 & 0.05 & 0.17 & 0.71 & 0.31 \\
UniAD & -&\checkmark& \checkmark & 0.20 & 0.42 & 0.75& 0.46 & 0.02 & 0.25 & 0.84&0.37  \\

VAD-Base~\cite{jiang2023vad}&- &-&-& 0.69 & 1.22 & 1.83 &1.25 & 0.06 & 0.68 & 2.52 &1.09  \\
VAD-Base&-&\checkmark&-& 0.41 & 0.70 & 1.06&0.72 & 0.04 & 0.43 & 1.15 &0.54\\
VAD-Base &-&\checkmark& \checkmark& 0.17 & 0.34 & 0.60 &0.37 & 0.04 & 0.27 & 0.67 & 0.33  \\
\midrule
Ego-MLP~\cite{zhai2023rethinking} &- & - & \checkmark& 0.15 & 0.32 & 0.59  & 0.35&0.00 & 0.27 & 0.85&0.37\\
BEV-Planner~\cite{li2024ego}&- &- &- & 0.30 & 0.52&0.83 &0.55 & 0.10 & 0.37 & 1.30 &0.59\\
BEV-Planner++ &- &\checkmark &\checkmark & 0.16 & 0.32& 0.57 & 0.35& 0.00 & 0.29 & 0.73 &0.34 \\

\midrule
DriveVLM\textdagger~\cite{tian2024drivevlm}& \checkmark &- & - &0.18 &0.34 &0.68 &0.40 &0.10 &0.22 &0.45 &0.27 \\
DriveVLM-Dual~\cite{tian2024drivevlm}& \checkmark &\checkmark &- & 0.15 & 0.29 & 0.48 & 0.31 & 0.05 & 0.08 & 0.17 & 0.10 \\

OmniDrive$\ddagger$~\cite{wang2024omnidrive}& \checkmark &-&-& 1.15 & 1.96 & 2.84 & 1.98 & 0.80 & 3.12 & 7.46 & 3.79 \\
OmniDrive & \checkmark &-&-& 0.40 & 0.80 & 1.32 & 0.84 & 0.04 & 0.46 & 2.32 & 0.94 \\
OmniDrive++& \checkmark &\checkmark &\checkmark & 0.14 & 0.29 & 0.55 & 0.33 & 0.00 & 0.13 & 0.78 & 0.30 \\

Senna~\cite{jiang2024senna}& \checkmark &- &- & 0.37 & 0.54 & 0.86 & 0.59 & 0.09 & 0.12 & 0.33 & 0.18 \\
Senna& \checkmark &\checkmark &\checkmark & \textbf{0.11} & \textbf{0.21} & \textbf{0.35} & \textbf{0.22} & 0.04 & \textbf{0.08} & \textbf{0.13} & \textbf{0.08}\\

EMMA\textdagger~\cite{hwang2024emma}& \checkmark &- &- &0.14 &0.29 &0.54 &0.32&- &- &- &-\\

\midrule
ORION (\textbf{Ours})& \checkmark &\checkmark & - & 0.17 & 0.31& 0.55 & 0.34& 0.05 & 0.25 & 0.80 & 0.37  \\
\bottomrule
\end{tabular}
\label{tab:sota-plan}
\end{table*}

\noindent We provide supplementary material to complement the main paper, arranged as follows:
\begin{itemize}
     \item Appendix~\ref{sup: ORION-b2d}: Details on the Chat-B2D dataset.
    \item Appendix~\ref{sup: traning_details}: Traning Details.
    \item Appendix~\ref{sup: more_result}: More results.
\end{itemize}

\section{Details on the Chat-B2D dataset}
\label{sup: ORION-b2d}
To compensate for the absence of a high-quality scene text annotation dataset and promote the application of VLM in the closed-loop simulated driving scenario, we carefully design an automated annotation pipeline to extend the Bench2Drive dataset~\cite{jia2024bench2drive} to support VQA pairs, named Chat-B2D, covering diverse tasks.

\subsection{Data Annotation Pipeline}
\label{sup: data annotation pipeline}
\begin{figure}[t!]
    \centering
    \includegraphics[width=0.47\textwidth]{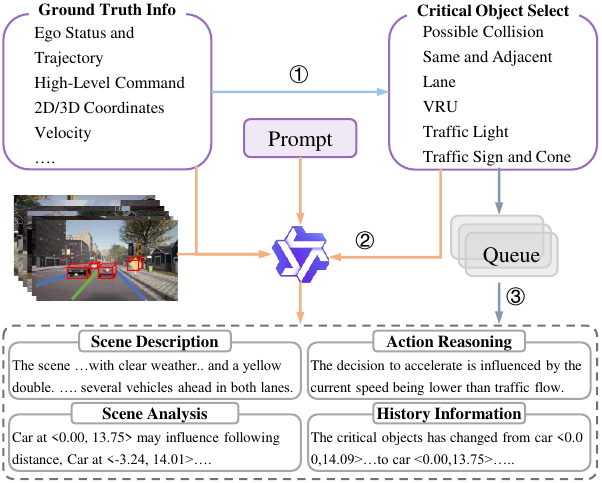}
    \caption{The automated annotation pipeline for the Chat-B2D dataset.}
    \label{fig: annotation}
\end{figure}

As shown in Fig.~\ref{fig: annotation}, the automated annotation pipeline consists of three steps:

\textbf{Critical object selection.} Unlike mainline self-driving perception modules that process all detected objects equally, we emphasize identifying the crucial object that potentially affects the ego vehicle’s driving behavior, grounded in human driving strategies. Our selection criteria include: 1) Objects have potential collisions within three seconds. 2) Leading vehicles in current and adjacent lanes. 3) Active traffic signals. 4) The vulnerable road users (VRUs), such as pedestrians/cyclists.

\textbf{Description generation.} We extract video clips comprising the current and five preceding frames. Subsequently, these clips, along with the ego vehicle's status and the ground truth information (\textit{e.g.}, 2D/3D coordinates and velocity, \textit{etc.}) of selected crucial objects, serve as input to Qwen2VL-72B~\cite{wang2024qwen2vl} for multi-task generation: 1) the scene description; 2) attributes of key objects and their impact on the ego vehicle; 3) operational meta-commands and action reasoning for autonomous navigation.

\textbf{History Information.} During the generation process, we incorporate a queue mechanism to preserve essential historical information. The stored information comprises two principal components: 1) Environmental dynamics that capture spatiotemporal variations of critical scene elements, and 2) Ego-motion characteristics derived from comparative analysis between current speed/action and their historical counterparts across previous frames.

The generated description and collected historical information are combined with predefined question templates to create VQA pairs. Tab.~\ref{tab: dataset prompt} displays the detailed crafted prompt, and Tab.~\ref{tab: question template} shows the question templates.

\subsection{Chat-B2D Attribute}
\label{sup: dataset attribute}
Through the carefully crafted prompts and the above generation pipeline, we have automatically conducted a large-scale, high-quality VQA dataset for the Bench2Drive~\cite{jia2024bench2drive}, creating Chat-B2D. This dataset, including a total of 2.11M VQA pairs for training and 0.12M  for validation, supports four primary categories: 1) Scene description, which provides a comprehensive overview of the driving scenarios, including weather, time of day, traffic situations, and road characteristics. 2) Behavior description of critical objects detailing their current state and intentions. 3) Meta-driving decisions and action reasoning of the ego car, such as turning left and lane following. 4) Recall of essential historical information.

\section{Training Details}
\label{sup: traning_details}

To accelerate the alignment of the vision-reasoning-action space and gradually enhance the reasoning and planning capabilities of our ORION, we adopt a three-stage training strategy. In each stage, the model inherits the weights from the previous stage and continues training. Additionally, we train the model for six epochs per stage with a total batch size of 32. The three-stage training strategy is as follows:

\noindent\textbf{3D Vision-Language Alignment:} In this first stage, we primarily train the QT-Former and the VLM while freezing the generative planner. By training on VQA pairs from Chat-B2D, we focus on aligning the vision space with the reasoning space.

\noindent\textbf{Language-Action Alignment:} In this stage, we unfreeze the generative planner and train the entire model except for the LLM, which is trained by LoRA~\cite{hu2021lora}, to predict planning trajectories without auxiliary VQA pairs. This stage primarily focuses on transmitting world knowledge from the reasoning space to the action space.

\noindent\textbf{End-to-End Fine-tuning:} We follow the training settings from the previous stage, with the only difference being the incorporation of joint training on VQA and planning tasks. This step further facilitates the alignment of the vision-reasoning-action space.

\section{More Results}
\label{sup: more_result}

\subsection{Experiments on nuScenes dataset}
\label{sup: nuScenes result}

\noindent \textbf{nuScenes Dataset.} nuScenes~\cite{caesar2020nuscenes} is a popular autonomous driving benchmark typically used for detection and open-loop planning evaluation. The dataset contains 1000 scenes from Singapore and Boston, with 700 scenes for training, 150 scenes for validation, and 150 scenes for testing. Each scene spans 20 seconds and is annotated at 2 Hz. nuScenes utilizes the L2 error and collision rate as planning metrics.

\noindent \textbf{Results on nuScenes.} We compare the ORION with previous SOTA end-to-end autonomous driving methods on the nuScenes dataset. Here, for a fair comparison with other VLM-Based methods, we modify ORION by replacing QT-Former with the Q-Former from OmniDrive~\cite{wang2024omnidrive}, and without the explicit ego status in the generative planner. As shown in Tab.~\ref{tab: nuscene results}, our ORION achieves comparable performance to classic SoTA methods~\cite{hu2023planning, jiang2023vad, li2024ego} without VLM. However, compared with other VLM-Based methods, our ORION is suboptimal. We argue that this is due to the latent space of VAE being more suitable for multimodal trajectory distributions of Bench2Drive~\cite{jia2024bench2drive}. In contrast, the nuScene dataset follows a uni-modal Gaussian distribution (with straight trajectories accounting for about 70\%). 

Additionally, as highlighted in BEV-Planner~\cite{li2024ego} and Ego-MLP~\cite{zhai2023rethinking}, even a simple MLP decoder with ego status can achieve strong open-loop planning performance on nuScenes. Thus, in the main paper, we primarily focus on evaluating ORION’s closed-loop performance on the Bench2Drive dataset.

\begin{figure}[t!]
    \centering
    \includegraphics[width=0.47\textwidth]{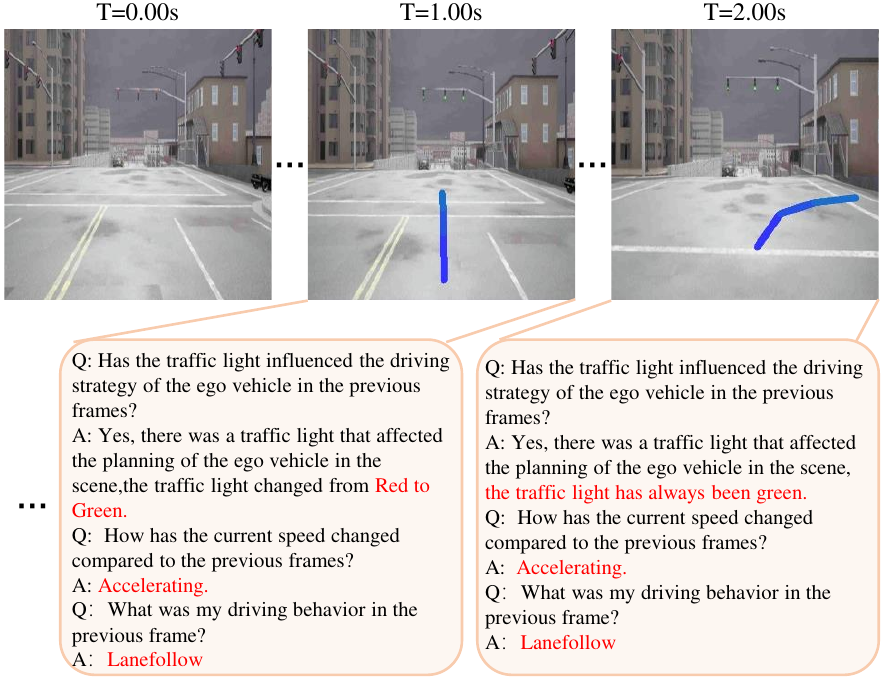}
    \caption{Qualitative results of historical information memory and retrieval on Bench2Drive open-loop validation set.}
    \label{fig: history vis}
\end{figure}

\subsection{More Ablation Studies on Bench2Drive}

\begin{table}[tp!]
\centering
\caption{Ablation study of training strategy. V/L/A indicates vision/language/action space. DS and SR denote Driving Score and Success Rate separately. C/B/R refers to CIDEr/BLEU/ROUGE-L.}
\label{tab: training pipeline ablation}
% \resizebox{1.0\textwidth}{!}{\begin{tabular}{l| c | c c c c}
\footnotesize
\setlength\tabcolsep{3.0mm} %列
\begin{tabular}{c  c c c   c c}
\toprule
\multirow{2.3}{*}{ID} & \multirow{2.3}{*}{V$\to$L} & \multirow{2.3}{*}{L$\to$A} & \multirow{2.3}{*}{V$\to$L$\to$A} & \multicolumn{2}{c}{Closed-loop}   \\ 
\cmidrule(lr){5-6}  & & &   & DS $\uparrow$  & SR(\%) $\uparrow$ \\ 
\midrule
1 &  & \checkmark  &      &57.96  &26.32 \\
2 &\checkmark   & \checkmark  &       & 65.10   &  38.83  \\
3 &\checkmark  & \checkmark  & \checkmark  &  74.65 & 49.31  \\ 

\bottomrule
\end{tabular}
\end{table}

\noindent\textbf{Ablation of training pipeline.} To facilitate the vision-language-action space alignment of our model,  we implement a progressive space alignment training strategy. We validate the effectiveness of the training pipeline, and the results are presented in Tab.~\ref{tab: training pipeline ablation}. Here, the QT-Former of our model does not incorporate collision loss or long-term memory bank with history queries. Specifically, through our second-stage training(ID-2), ORION achieves a significant improvement by +7.14 DS and +12.51\% SR compared to direct training planning without the first stage (ID-1). After completing the third-stage training (ID-3), our model further improved the performance and achieved optimal (74.65 DS and 49.32 SR), demonstrating the effectiveness of our training strategy.

\subsection{More Qualitative Results}
\label{sup: visualization}
\noindent\textbf{Historical information memory and retrieval.} Benefiting from the introduced long-term memory bank and history queries in QT-Former,  our ORION could store and retrieve historical information, as illustrated in Fig.~\ref{fig: history vis}. Our model could perceive critical elements (\textit{e.g.}, traffic light) changes in previous and current times.

\begin{figure*}[htbp]
    \centering
    \includegraphics[width=0.9\textwidth]{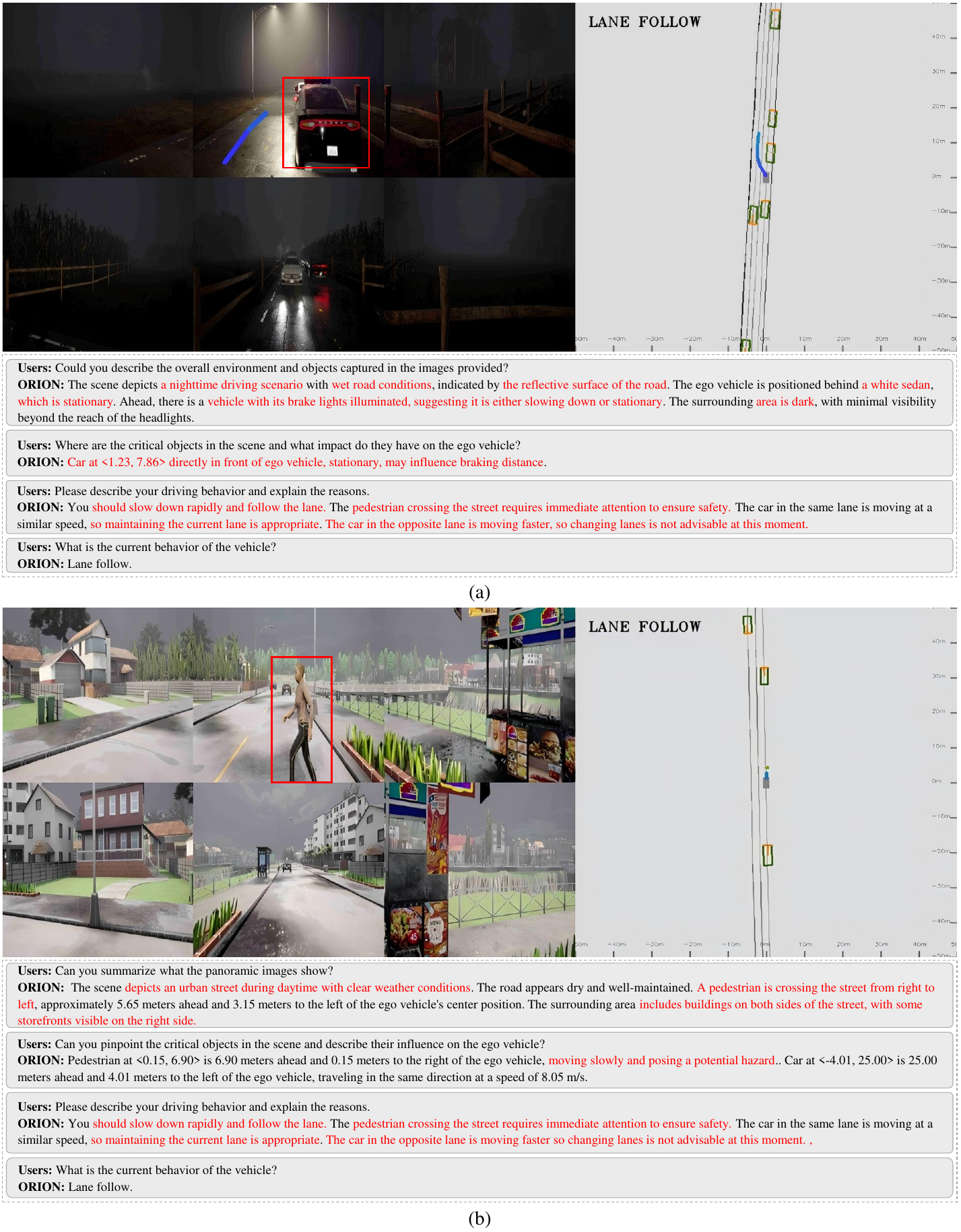}
    \caption{Qualitative results for scene understanding and action reasoning on Bench2Drive open-loop validation. From top to bottom, each sub-figure displays the multi-view input and traffic conditions in Bird's Eye View (BEV)  of the current scene, the scene understanding, and the reasoning result. The red rectangles indicate the critical objects influencing the action of the ego vehicle,  while the red text highlights our method’s correct scene comprehension.}
    \label{fig: sup_vis}
\end{figure*}

\noindent\textbf{Scene understanding and action reasoning.} Fig.~\ref{fig: sup_vis} shows scene understanding and action reasoning results of ORION. It could be observed that ORION could not only accurately perceive detailed scene information but also identify key objects influencing the ego vehicle’s behavior and infer appropriate motion decisions. Even in extreme situations (e.g., a pedestrian suddenly crosses the road in Fig.~\ref{fig: sup_vis}(b)), our model maintains robust performance, highlighting its superior reasoning and decision-making ability.

\begin{table*}[t!]
\centering
\caption{Prompts fed into Qwen2VL to generate corresponding response.}
\label{tab: dataset prompt}
\begin{tcolorbox}[colback=gray!10]
\centering
\footnotesize
\begin{tabular}{p{0.97\columnwidth} c}

\PromptSty{\textbf{Prompt 1: Scene Description}}\\

Suppose you are driving, generate a description of the driving scene which includes the key factors for driving planning, including the traffic conditions, weather, time of day and road conditions, traffic signs, and traffic lights that affect the driving of the ego vehicle if it exists, indicating smooth surfaces or the presence of obstacles; The description should be concise, and accurate to facilitate informed decision-making. Please make sure the traffic light colors you provide are accurate; otherwise, give `unknown.' \\

\hrulefill \\ 

\PromptSty{\textbf{Prompt 2: Critical Objects Analysis}}\\
I will provide you with several critical objects that are most important to my short-term command in the last image of the video. 
I provide you with 2d coordinates, which are two points of the top-left and bottom-right coordinates, and the 3d position and velocity information of these critical objects: \{objects\_desc\}. Please describe their action and explain why they are most important, including their speed, position, heading, and influence on ego vehicle. Please associate these objects with the objects in the image and please remember the ego vehicle is located at the **center of the bottom edge of the picture**. \\

\hrulefill \\ 

\PromptSty{\textbf{Prompt 3: Expert Meta-Decision}}\\

Besides, I will provide you speed, historical trajectory and future driving behaviors of ego vehicle, which can be divided into SPEED decisions and COMMAND decisions, SPEED includes keep, accelerate, decelerate, while COMMAND includes left, right, straight, lane follow, change lane left, change lane right. Your current speed is \{ego\_vel\} m/s, historical trajectory is \{ego\_his\_trajs\}. The next SPEED decision is \{speed\_decision\}, the next COMMAND decision is \{path\_decision\}. Please analyze the reasons for the future driving behaviors or the reason why ego vehicle can \{path\_decision\} based on the driving environment, including the behavior of other traffic participants, especially the critical objects, road conditions, and traffic light status. \\

\hrulefill \\ 

\PromptSty{\textbf{Example: }}\\

You should refer to the following example and format the results like \{``description": ``xxx",``critical\_objects": ``xxx", ``action": ``\{speed\_decision\} and \{path\_decision\}"\}\}:

\{\{ ``description": ``The scene captures a moment of urban life framed by a red traffic light in mid-transition. To the right, a pedestrian crossing, ..., waiting for the signal to change. 
Directly ahead, ...
On the left, the sidewalk bustles with people of all ages, ... 
Behind this foreground of orderly traffic and pedestrian movement, ..." \\

``critical\_objects: ``[``Car at $<$-0.24, 7.56$>$ directly in front of ego vehicle, ...", ``Car at $<$-2.64, 10.00$>$ ..., moving at a slower speed, may influence left change.``]" \\

``action": ``Slow down and right lane change. 
- The decision to change lanes is influenced by the need to overtake Car at $<$-0.24, 7.56$>$ in front of the ego vehicle. 
- There are no traffic lights for the vehicle,... 
- Pedestrians are visible on the sidewalk to the right, ..."
\}\} \\

If it has no critical\_objects, you should refer to the following example and format the results like \{\{``description": ``xxx", ``critical\_objects": [], ``action": "xxx"\}\}.\\
\end{tabular}
\end{tcolorbox}
% \captionof{figure}{\textbf{Prompts fed into Qwen2VL to generate corresponding response.}}
% \label{fig: dataset prompt}
\end{table*}
\begin{table*}[t!]
\centering
\caption{A list of question templates for diverse VQA tasks.}
\label{tab: question template}
\begin{tcolorbox}[colback=gray!10]
\centering
\footnotesize
\begin{tabular}{p{0.97\columnwidth} c}

\PromptSty{\textbf{Type 1: Scene Description}}\\
1. What can you tell about the current driving conditions from the images? \\
2. What can be observed in the panoramic images provided? \\
3. Can you provide a summary of the current driving scenario based on the input images? \\
4. What can you observe from the provided images regarding the driving conditions? \\
5. Please describe the current driving conditions based on the images provided. \\
6. Can you describe the current weather conditions and the general environment depicted in the images? \\
7. Please describe the current driving conditions based on the input images. \\
8. Could you summarize the current driving conditions based on the input images? \\
9. Please provide an overview of the current driving conditions based on the images. \\
10. Can you summarize what the panoramic images show? \\
11. Can you describe the overall conditions and environment based on the images? \\
12. Could you describe the overall environment and objects captured in the images provided? \\
\hrulefill \\ 

\PromptSty{\textbf{Type 2: Critical Objects Analysis}} \\
1. Where are the critical objects in the scene and what impact do they have on the ego vehicle? \\
2. Identify the significant objects in the scene and their specific impacts on the ego vehicle. \\
3. Can you pinpoint the critical objects in the scene and describe their influence on the ego vehicle? \\
4. Which objects in the scene are critical, and what effects do they have on the ego vehicle's movement? \\
5. Please describe the critical objects in the scene, their positions, and the influence they have on the ego vehicle. \\
\hrulefill \\ 

\PromptSty{\textbf{Type 3: Interpretable Action of Ego Vehicle}} \\
1. Please describe your driving behavior and explain the reasons. \\
2. What is the current behavior of the vehicle? \\
\hrulefill \\ 

\PromptSty{\textbf{Type 4: Historical Information}}\\
1. What are the differences between the current scene and the past scene in terms of critical objects? \\
2. How do the critical objects in the current scene differ from those in the past scene? \\
3. What changes have occurred in the critical objects between the current and past scenes? \\
4. What are the differences in critical objects between the present scene and the previous scene? \\
5. What distinctions exist between the critical objects of the current scene and those of the past scene? \\

6. In the past few frames, has a traffic light affected the driving strategy of the ego vehicle? \\
7. Within the recent frames, has the driving strategy of the ego vehicle been influenced by a traffic light? \\
8. In the last few frames, has the driving strategy of the ego vehicle been impacted by a traffic light? \\
9. Has the driving strategy of the ego vehicle been affected by a traffic light in the past few frames? \\
10. Has the traffic light influenced the driving strategy of the ego vehicle in the previous frames? \\

11. How has the current speed changed compared to the previous frames? \\
12. What was my driving behavior in the previous frame? \\

\end{tabular}
\end{tcolorbox}
% \captionof{figure}{\textbf{A list of questions for diverse tasks.}}
% \label{fig: question template}
\end{table*}

\end{document}